%% file: coling2020.tex
\documentclass[11pt,a4paper]{article}
\usepackage{coling2020}
\usepackage{times}
\usepackage{latexsym}
\usepackage{graphicx}
\usepackage{todonotes}
\usepackage{booktabs}
\usepackage{amsmath}
\usepackage{amssymb}
\usepackage{tikz}
\usepackage{enumitem}
\usepackage{caption}
\usetikzlibrary{shapes}
\usepackage{url}
\usepackage{subfig}
\usetikzlibrary{calc,arrows.meta,positioning,arrows}
\usetikzlibrary{decorations.pathmorphing,shapes}

\usepackage{microtype}

\newcounter{sarrow}

\newcommand{\myrightleftarrows}[1]{\mathrel{\substack{\xrightarrow{#1} \\[-.9ex] \xleftarrow{#1}}}}

\colingfinalcopy

\title{Vec2Sent: Probing Sentence Embeddings with\\ Natural Language Generation}

\author{Martin Kerscher {\normalfont and} Steffen Eger \\
  Natural Language Learning Group (NLLG) \\
  Technische Universität Darmstadt \\
  \texttt{martin.kerscher@protonmail.com}, \texttt{eger@aiphes.tu-darmstadt.de}
}

\date{}

\begin{document}
\maketitle
\begin{abstract}
We introspect 
black-box sentence embeddings by conditionally generating from them with the objective to retrieve the underlying discrete sentence. We perceive of 
this 
as a new unsupervised probing task and show that it correlates well with downstream task performance. We also illustrate how 
the language generated from different encoders differs. 
We apply our approach to generate sentence analogies from sentence embeddings. 
\end{abstract}

\input{introduction.tex}
\input{approach.tex}
\input{experiments.tex}
\input{conclusion.tex}

\bibliography{bibliography_2020}
\bibliographystyle{coling}

\appendix
\newpage
\section{Supplementary Material}
\input{myappendix.tex}

\end{document}

%% file: introduction.tex
\section{Introduction}
Generalizing the concept of word embeddings to 
sentence level, sentence embeddings (a.k.a.\ sentence encoders) are 
ubiquitous in NLP as features in downstream classification tasks and in semantic similarity and retrieval applications \cite{Kiros.2015,Conneau.2017}. 
\emph{Probing sentence encoders} for the linguistic information signals they contain has likewise become an important field of research, as this 
allows to introspect otherwise black-box representations \cite{Adi.2017,Conneau.2018a}. 
The idea behind probing tasks is to 
query representations for certain kinds of linguistic information such as the dependency tree depth of an encoded sentence. There are a variety of 
problems surrounding current probing task specifications: (i) probing tasks need to be manually construed, which 
brings with it a certain degree of arbitrariness and incompleteness; (ii) most probing tasks require labeled datasets or 
trained classifiers such as dependency parsers for linguistic processing---however, these may be unavailable for many low-resource languages or available only to a limited degree; (iii) it is not entirely clear how probing tasks have to be designed, 
e.g., how much training data they require and which classifier to use for probing \cite{Eger_howto:2020}; (iv) 
\newcite{Ravichander2020ProbingTP} also argue that standard probing tasks do not outline the information signals a classifier actually uses for making predictions.

\textbf{Our contribution} 
is 
an alternative, more direct introspection of sentence embeddings, namely, through conditional natural language generation, which we call ``vec2sent'' (V2S). 
By retrieving and (manually) investigating the discrete output obtained from a dense vector representation, linguistic properties of the embedding may be `directly' unveiled: e.g., we expect that a word-order insensitive model 
would have a comparatively hard time in restoring the correct syntax of an encoded sentence. 
V2S requires no labeled data, 
making it applicable to any language that has at least several ten thousands of written sentences available---e.g., it is particularly suitable for multilingual probing \cite{Krasnowska.2019,Eger_howto:2020}. 
Since V2S makes the opaque space $\mathbb{R}^d$ observable, it may also reveal intriguing properties of how encoders encode text (cf.\ Table \ref{table:examples}), without having to `guess' relevant probing tasks.\footnote{Our code is available at \url{https://github.com/maruker/vec2sent}.}

%% file: approach.tex
\section{Approach}
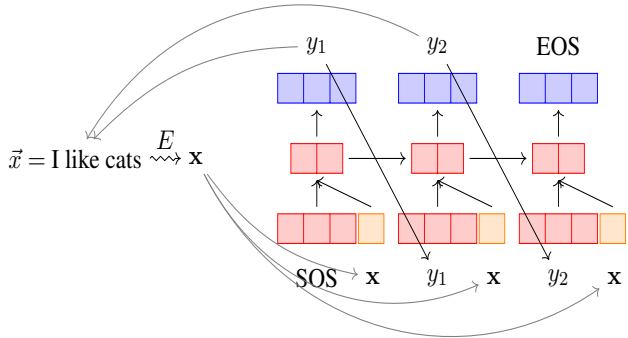
\begin{figure}[!htb]
\centering
\resizebox{8.5cm}{5.5cm}{
\input{figures/model.tex}
}
\caption{Schematic illustration of our approach: A  pre-trained sentence encoder $E$ maps the input $\vec{x}$ to a vector $\mathbf{x}$. This vector is fed into an RNN language model. 
The output $\vec{y}=(y_1,y_2,\ldots)$ generated by the conditional RNN is finally compared to the original input sentence.} 
\label{fig:model}
\end{figure}

For an input sentence $\vec{x}\in V^*$ (a sequence of tokens over a vocabulary $V$), consider its sentence representation $\mathbf{x}\in\mathbb{R}^d$ induced by some sentence embedding model $E$. 
We consider a 
decoder $D$ which takes $\mathbf{x}$ as input and produces natural language text as output. We train $D$ to reconstruct $\vec{x}$ from $\mathbf{x}$. 
For simplicity, we design $D$ as a recurrent neural network (RNN) with LSTM cells, rather than 
as a more recent model class like Transformers. 
At each time step $i$ in the RNN, the goal is to predict the next word given the previously generated word $y_{i-1}$ as input as well as the hidden state vector $\mathbf{h}_i$ which summarizes all past observations.  
To implement generation conditional upon the sentence embedding $\mathbf{x}$, we 
concatenate $\mathbf{x}$ to each input word embedding   $\mathbf{y}_{i-1}$. 
Figure \ref{fig:model} 
illustrates our approach.  

In general, V2S translates from a continuous vector space to a discrete language space, while an embedding model performs the inverse operation:
\begin{align*}
    \mathbf{x}\in\mathbb{R}^d\quad \mathrel{\substack{\text{V2S}\\ 
    \myrightleftarrows{\rule{1cm}{0cm}}
    \\ E}}\quad \vec{x}\in V^*
\end{align*}
The continuous space in which $\mathbf{x}$ lies is `opaque' (to humans) and requires probing tasks for introspection. The discrete space in which $\vec{x}$ lies can easily be introspected by humans (at least at small scale). 
An interesting application of the duality of $E$ and V2S is that one can interpolate any two vectors $\mathbf{x}$ and $\mathbf{y}$ via 
$\mathbf{z}(\alpha)=\alpha\cdot \mathbf{x}+(1-\alpha)\mathbf{y}$---using V2S, one can then decode $\mathbf{z}(\alpha)$ to derive its discrete representation for all sentences $\vec{z}(\alpha)$ ``between'' $\vec{x}$ and $\vec{y}$. 
Analogously, we could decode sentence representations 
to find analogies akin the famous king-woman+man equation for word embeddings \cite{diallo-etal-2019-learning}.  

\paragraph{Diagnostic tests} The output $\vec{y}$ of the language model can be compared to the original input 
$\vec{x}$. Intuitively, the closer $\vec{y}$ is to $\vec{x}$,  
the better. While manual introspection of the results may yield important insights into the deficiencies (or capabilities) of an encoder, human introspection may both be unfeasible at large scale and subjective. To this end, we define several diagnostic tests: 
\begin{itemize}[topsep=3pt,itemsep=-1pt,leftmargin=*]
    \item \texttt{Id}$(\vec{x},\vec{y})$: the fraction of sentences for which the condition $\vec{y}=\vec{x}$ holds.
    \item \texttt{PERM}$(\vec{x},\vec{y})$: the fraction of sentences where $\vec{y}$ is a permutation of $\vec{x}$, i.e., whether $\vec{y}$ can be obtained by permuting the words in $\vec{x}$.
    \item \texttt{Id}/\texttt{PERM}: the division of \texttt{Id} by \texttt{PERM}.
    \item \texttt{BLEU}$(\vec{x},\vec{y})$: measures the n-gram overlap between $\vec{x}$ and $\vec{y}$. We report average BLEU across our sample of sentences.
    \item \texttt{Mover}$(\vec{x},\vec{y})$: 
    We use MoverScore \cite{zhao2019moverscore}, a recent 
    soft evaluation metric based on BERT embeddings \cite{Devlin.2018}, which has shown much better  correlations with humans than BLEU. 
\end{itemize}

%% file: figures/model.tex
\begin{tikzpicture}[
  hid/.style 2 args={
    rectangle split,
    rectangle split horizontal,
    draw=#2,
    rectangle split parts=#1,
    fill=#2!20,
    outer sep=1mm}]
  \node (mytext) at (-2,0) {$\vec{x}$ $=$ I like cats};
  \node (emb) at (0,0) {$\mathbf{x}$};
  \foreach \i [count=\step from 1] in {SOS,$y_1$,$y_2$} 
    \node (i\step) at (2*\step-0., -2+0.3) {\i};
    
 \foreach \i [count=\step from 1] in {$\mathbf{x}$,$\mathbf{x}$,$\mathbf{x}$} 
    \node (ii\step) at (2*\step+0.93, -2+0.3) {\i};
  \foreach \step in {1,...,3} {
    \node[hid={2}{red}] (h\step) at (2*\step, 0) {};
    \node[hid={3}{red}] (e\step) at (2*\step, -1) {}; 
    \node[hid={3}{blue}] (f\step) at (2*\step, 1) {};
    \node[hid={1}{orange}] (q\step) at (2*\step+0.9, -1) {};
    \draw[->] (e\step.north) -> (h\step.south);
    \draw[->] (h\step.north) -> (f\step.south);
    \draw[->] (q\step.north) -> (h\step.south);
  }
  \draw [->,
line join=round,
decorate, decoration={
    zigzag,
    segment length=4,
    amplitude=.9,post=lineto,
    post length=2pt
}]  (mytext.east) -- node[above]{$E$} (emb.west);
  \draw[->,gray] (emb) to [bend right=20] (ii1);
  \draw[->,gray] (emb) to [bend right=40] (ii2);
  \draw[->,gray] (emb) to [bend right=45] (ii3);

  \foreach \step in {1,...,2} {
    \pgfmathtruncatemacro{\next}{add(\step,1)}
    \draw[->] (h\step.east) -> (h\next.west);
  }
  
  \node (h1) at (2, +1.6) {$y_1$};
  \node (h2) at (4, +1.6) {$y_2$};
  \node (h4) at (6, +1.6) {EOS};
  \draw[->] (h1) -> (i2); 
  \draw[->] (h2) -> (i3);
  
    \draw[->,gray] (h1) to [bend right=20] (mytext);
    \draw[->,gray] (h2) to [bend right=40] (mytext);

\end{tikzpicture}

%% file: experiments.tex
\section{Experiments} \label{sec:experiments}

\paragraph{Sentence Encoders}
We consider two types of sentence encoders, \emph{non-parametric methods} which combine word embeddings in elementary ways, without training; and \emph{parametric methods}, which tune parameters on top of word embeddings. 
As \textbf{non-parametric methods}, we consider: (i) average word embeddings as a 
popular baseline, (ii) GEM \cite{yang2018breaking}, a 
weighted averaging model, 
(iii) hierarchical embeddings \cite{shen-etal-2018-baseline}, an order-sensitive model where a max-pooling operation is applied to averages of word 3-gram embeddings in a sentence, (iv) the concatenation of average, hierarchical and max pooling \cite{Rueckle.2018}, 
and (v) sent2vec \cite{Pagliardini.2018}, 
a compositional word n-gram model. 
For (i)-(iv) we use BPEmb subword embeddings \cite{heinzerling-strube-2018-bpemb} as token representations. 

As \textbf{parametric methods}, we consider: (vi) InferSent \cite{Conneau.2017}, which induces a sentence representation by learning a semantic entailment relationship between two sentences; (vii) QuickThought \cite{Logeswaran.2018}  
which reframes 
the popular SkipThought model \cite{Kiros.2015} in a classification context; 
(viii) LASER \cite{Artetxe.2018} derived from massively multilingual machine translation models; and (ix) 
sentence BERT (SBERT) \cite{Reimers.2019}, which fine-tunes BERT representations 
on SNLI 
and then averages fine-tuned token embeddings to obtain a sentence representation.  
The encoders and their sizes are listed in Table \ref{tab:top3}. 

 \begin{minipage}{\textwidth}
  \begin{minipage}[b]{0.46\textwidth}
     \input{figures/tables/top3_main}
     \captionof{table}{Encoders, their dimensionalities, and rank of encoders according to downstream (DS) and V2S diagnostic tasks. Ranks for DS are 
     after averaging across all tasks; ranks for V2S are from the \texttt{Id} diagnostic test.}
     \label{tab:top3}
    \end{minipage}
  \hspace{.199cm}
  \begin{minipage}[b]{0.46\textwidth}
    \centering
    \includegraphics[scale=0.185]{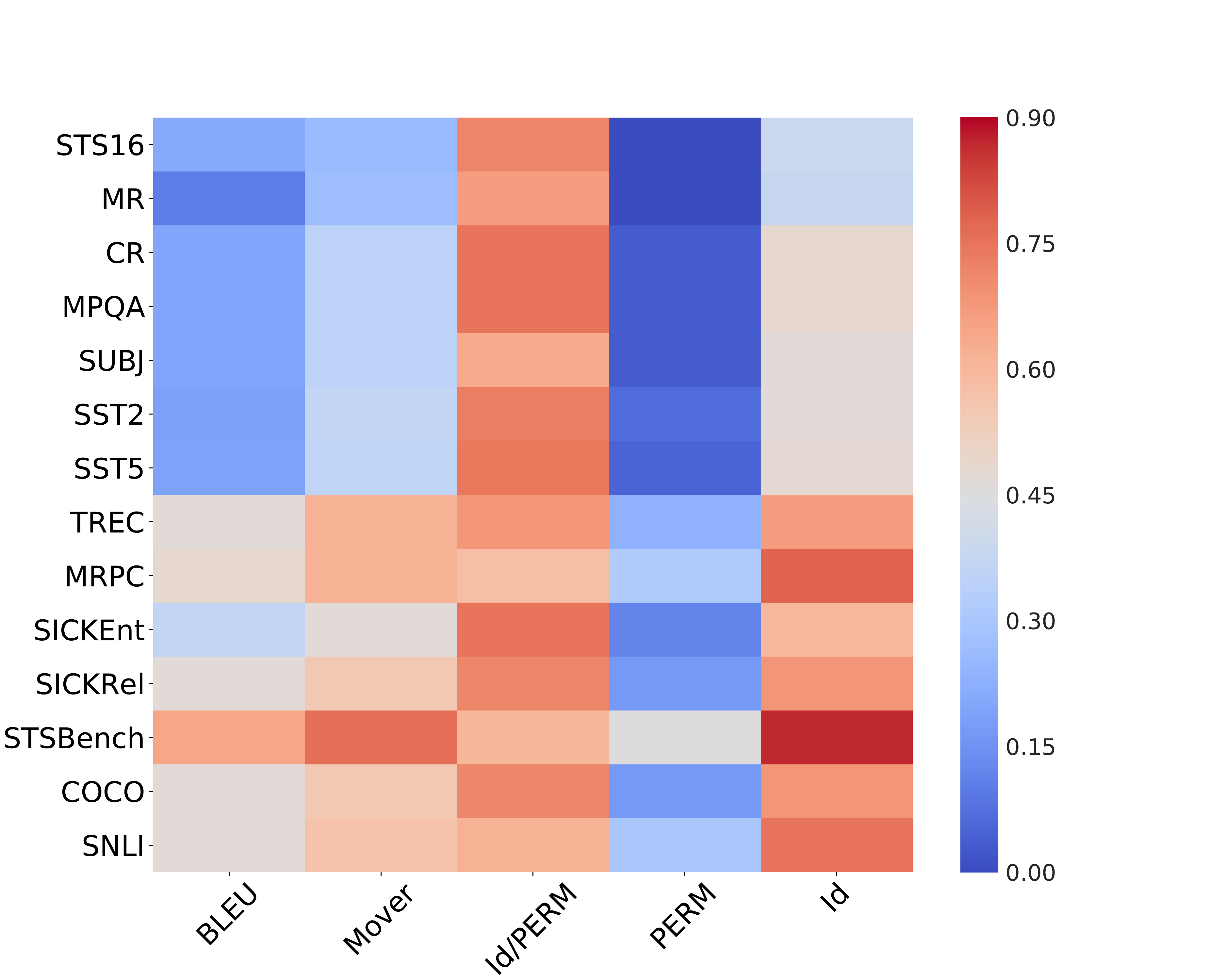}
    \captionof{figure}{Spearman rank correlation betweendownstream tasks from SentEval and V2S diagnostic tests.}
    \label{fig:rank_spearman}
  \end{minipage}
\end{minipage}
\vspace{0.15cm}

\paragraph{Setup} As a decoder $D$, we use the Mixture of Softmax RNN language model (MOS) proposed in \newcite{yang2018breaking}, which addresses the `softmax bottleneck' in standard softmax language models and has achieved excellent results on benchmark tasks. For each encoder, we train on 1.4m  
sentences from  
News Crawl 2007\footnote{http://www.statmt.org/wmt14/training-monolingual-news-crawl/news.2007.en.shuffled.gz}
with the objective to retrieve the original sentence $\vec{x}$ from $\mathbf{x}=E(\vec{x})$. 
We evaluate performance on 10k sentences from News Crawl 2008. 
For speed reasons, we 
restrict sentence length to at most 15 tokens. 
Our RNN has 3 hidden layers with 1024 hidden units in each. In \S\ref{sec:stability}, we verify that the design choice of the decoder (e.g., MOS architecture vs.\ a simple RNN) plays a marginal role for the conclusions of V2S in our experiments.

In the sequel, we test whether V2S can predict downstream task performance (\S\ref{sec:down}) before introspecting the language generated by different encoders (\S\ref{sec:analysis}). 

\subsection{Correlation with Downstream Tasks}\label{sec:down}
In Figure  
\ref{fig:rank_spearman}, 
we plot Spearman rank correlations (as in \newcite{Conneau.2018b}) between 
our V2S statistics and the 14 downstream tasks for standardized comparison of sentence encoders from SentEval \cite{Conneau.2018b}; full results are given in the appendix. 
The correlations indicate whether rankings of encoders induced by the probing tasks 
transfer to downstream tasks: i.e., to which degree it holds that $A\succ_{\text{Probing}} B$ implies that $A\succ_{\text{Downstream}} B$, for encoders $A$ and $B$. 
The downstream tasks and V2S have no training data overlap. 

\paragraph{Results} 
We observe the following: (i) \texttt{PERM} is not strongly (but in general positively) correlated with downstream task performance. This means that an encoder which can better identify all the words of $\vec{x}$ from $\mathbf{x}$ has (only) a small chance of performing better in a downstream task---in fact, Avg has the best performance for \texttt{PERM} but typically performs badly in downstream tasks. 
(ii)
\texttt{Mover} beats \texttt{BLEU}.
This is as expected, as \texttt{Mover} can better take paraphrasing and lexical variation into account. (iii) More surprising is that \texttt{Id} beats both of them, even though \texttt{Id} is one of the simplest possible metrics: it only asks whether $\vec{x}$ could exactly be reproduced from $\mathbf{x}$. 
(iv) The best correlation with downstream tasks in our context has \texttt{Id}/\texttt{PERM}. This means that an encoder performs better on downstream tasks if it satisfies two conditions: it can correctly identify all words in $\vec{x}$ \emph{and} place them in correct word order. 

\paragraph{Comparison with (other) probing tasks}
We contrast with the other 10 probing tasks defined in SentEval.  Compared to the likewise unsupervised Word Content (WC) tasks---which probes for specific lexical words stored in the representations---\texttt{Id} and \texttt{Id}/\texttt{PERM}
perform much better. 
In fact, WC has lowest predictiveness in our experiments.\footnote{While this is in contrast with \newcite{Conneau.2018a}, who report best predictiveness of WC, it is more congruent with \newcite{Perone.2018} and \newcite{Eger.2019}, who report low discrimination capabilities of WC. In the end, the choice of encoders may also play an important role in this context.} 
For instance, the average correlation with downstream tasks of WC is 0.23 and the lowest correlation is -0.09, while 
\texttt{Id} has an average correlation of 0.59 and a minimum of 0.38. 
\texttt{Id}/\texttt{PERM} has an average correlation of 0.69 and a minimum  of 0.58. 
Overall, \texttt{Id}/\texttt{PERM} is a better predictor of downstream task performance than 7 out of 10 probing task from SentEval according to the `min' category. 
It is beaten only by the three syntactic tasks TreeDepth, TopConstituents and BigramShift. When we remove SBERT as an encoder, \texttt{Id} has the best predictiveness according to 
`min' among all probing tasks and diagnostic tests. 
Altogether, this shows that statistics based on V2S can be excellent predictors for downstream task performance 
for our class of sentence encoders. 

It is unclear why SBERT performs rather badly according to most V2S statistics but has strong downstream task performance; cf.\ the rank statistics in Table  \ref{tab:top3}.\footnote{We remark that \newcite{Lin:2019} also report rather low performance for BERT based encoders according to the nearest neighbor overlap probing task.} 
Possibly, reconstructing text from SBERT 
representations is more difficult due to the contextualized nature of the embeddings and the SNLI fine-tuning objective \cite{SBERT-WK}. Note, however, that 
\texttt{Id}/\texttt{Perm} 
ranks SBERT on place 2nd, much closer to its actual downstream task performance, where it is overall the best encoder. 

\subsection{Qualitative Analysis}\label{sec:analysis}
\input{figures/tables/example.tex}
Table \ref{table:examples} gives a sample input sentence $\vec{x}$ and the content retrieved from its embedding $\mathbf{x}$, for various encoders. We see that  
LASER and InferSent embeddings almost completely contain the original sentence, in correct order; only the proper name (``Stoller'') and punctuation are not properly decoded. The Avg encoder gets almost all words correctly (except for missing a negation) but reverses their order, thus completely changing the text's meaning. 
The decoded sentence underlying the SBERT representation is 
furthest away from the surface form of the input, but semantically closer than, e.g., the Avg embedding. 

This is a vivid illustration why, e.g., the WC probing task may have little predictive power for downstream applications (contrary to what has been claimed in some previous research), as it may be better to represent the meaning and logical structure of a text rather than its surface form. We note that our examples are 
to some degree representative of the encoders. 
For example, among all encoders, Avg embeddings 
most frequently 
permute the word order of the input and 
we observe absence of correctly restored lexical information in many examples for SBERT. 

Next, we 
explore whether we can use V2S to induce sentence level semantic analogies. For three query sentences with representations $\mathbf{r}, \mathbf{s}, \mathbf{v}$ we compute the vector $\mathbf{u}:=\mathbf{r}-\mathbf{s}+\mathbf{v}$ (inspired by ``king-woman+man'') and give $\mathbf{u}$ as input to V2S. Illustrative results are shown in Table 
\ref{table:examples} (right). 
In many instances, 
we find that Avg performs poorly and cannot solve our semantic analogies. QT, SBERT, and LASER perform much better. LASER appears to encode gender biases in the shown example. SBERT, again, seemingly stores less surface level information. Note that it is in principle also possible to look for nearest neighbors of $\mathbf{u}$ to check for semantic analogies; however, nearest neighbors depend on the sample of sentences available in the corpus---V2S can be seen as a more direct form of introspection.

\subsection{Stability of the analysis}\label{sec:stability}
Finally, we test whether the obtained results are stable under a change of architecture (replacing the MOS language model with a simple LSTM language model---which in addition  maps the embedding in the initial hidden state of the RNN via a linear transformation rather than concatenating to word embeddings---and identical training sizes) and training size for the language model (200k sentences instead of 1.4m for MOS). Table \ref{table:stability} shows the results. 
\input{figures/tables/correlations.tex}
We find that all statistics are quite stable ($\rho\ge .80$ for all diagnostic tests) and do not seem to vary much along the investigated two dimensions. This is a reassuring result, as it indicates that 
our results are not an artefact of any of these two choices. 

%% file: figures/tables/top3_main.tex
\centering
    {\footnotesize
    \begin{tabular}{l|crr} \toprule
         \textbf{Encoder} & \textbf{Size} & 
         \textbf{Rank DS} & \textbf{Rank V2S} \\
         \midrule 
         Avg & 300 & 7 & 5 \\
         GEM &  300 &  9 & 9 \\
         Hier & 300 & 8 &  6 \\
         Avg$+$Max$+$Hier & 900 & 6 & 4\\
         Sent2Vec & 700 & 5 & 8 \\ \midrule
         InferSent & 4096 & 3 & 1\\
         QuickThought & 4800 & 2 & 2 \\
         LASER & 1024 & 4 & 3 \\
         SBERT & 1024 & 1 & 7 \\
         \bottomrule
    \end{tabular}
    }

%% file: figures/tables/example.tex
\begin{table}[!th]
    \centering
    {
    \begin{tabular}{p{1.3cm}|l|ll}
         \toprule 
         \textbf{Input} &  " the point is , " stoller adds , " i 'm not sure . & $a:b\,\,::\,\, \mathbf{z}:c$ & $a:b\,\,::\,\, \mathbf{z}:c$ \\ \midrule
         LASER &  " the point is , "inger adds , " i 'm not sure . & she visits italy & she is a nurse\\
         QT & " the point is , " i 'm not sure , " spector adds . & she visits france & she is a doctor\\
         InferSent &  the point is , bernstein adds , " i 'm not sure . " & rose visits italy & she is a doctor\\
         Avg & " i 'm sure , it is the point , " stoller adds . " & rate is & a.i.\\
         SBERT & " well , i don 't know , it 's my question , " steinbre & she travels spain & she is medicine\\
         \bottomrule
    \end{tabular}
    }
    \caption{Left: Input sentence $\vec{x}$, encoders and their sample reproductions $\vec{y}$ from vec2sent. Right: Sentence level analogies $a:b\,\,::\,\, z:c$ for the triples ($a=$ he visits italy, $b=$ he eats pizza, $c=$  she drinks wine) on the left, and ($a=$ he is a doctor, $b=$ his name is robert, $c=$ her name is julia) on the right. The notation $a:b\,\,::\,\, z:c$ means ``$a$ is to $b$ as $z$ to $c$''. We use V2S to induce the missing sentence $z$ in the analogy.}
    \label{table:examples}
\end{table}

%% file: figures/tables/correlations.tex
\begin{table}[!htb]
    \centering
    \begin{tabular}{l|rr} \toprule
         & Low-Res.\ MOS & Simple LSTM \\ \midrule
         \texttt{BLEU}& .90 & .87 \\
         \texttt{MOVER}& .97 & .95 \\
         \texttt{ID}& .92 & .90 \\
         \texttt{ID}/\texttt{PERM} & .90 & .82 \\
         \texttt{PERM}& .88 & .80 \\
         \bottomrule
    \end{tabular}
    \caption{Spearman rank correlation between ``high-resource'' MOS setting with 1.4m training sentences and (a) low-resource setting with same architecture (b) different architecture but same training size.}
    \label{table:stability}
\end{table}

%% file: conclusion.tex
\section{Discussion \& Conclusion}
The goal of probing tasks is to make the opaque embedding space $\mathbb{R}^d$ ``linguistically'' observable to humans. A plethora of different probing tasks have been suggested 
\cite{Linzen.2016,Shi.2016,Adi.2017},  
which \newcite{Conneau.2018a} classify into \emph{surface}, \emph{syntactic}, and \emph{semantic} probing tasks. 
More recently,  
multilingual extensions of probing tasks \cite{Krasnowska.2019,Sahin.2019,Eger_howto:2020} have been considered, as well as word level probing for especially contextualized representations  \cite{Tenney.2019,liu-etal-2019-linguistic}. In the latter context, \newcite{hewitt-manning-2019-structural} discover an intriguing structural property of BERT, \emph{viz.}, to contain whole syntax trees in its representations, possibly, as we show, at the expense of lexical information (at least SBERT). 

We investigated V2S as an alternative, direct way of probing sentence encoders. We showed that V2S may be a good predictor of downstream task performance and, in particular, that one of two simple diagnostic tests had good predictive performance in all scenarios: 
whether an encoder can exactly retrieve the underlying sentence from an embedding (\texttt{Id}) and whether the fraction of exactly retrieved sentences among all retrieved sentences that are permutations of the input sentence is high. 
Thus, 
we recommend to 
report both of these diagnostic tests for V2S. We also showed that V2S allows to directly introspect for certain structural properties of the embedding space, e.g., 
sentence-level semantic analogies. 

V2S is an unsupervised probing task and, as such,  
we believe that a particularly interesting use case will be for low-resource languages and scenarios, 
for which our experiments suggests that it will be a much better predictor of downstream task performance 
than the equally unsupervised (and most closely related) WC probing task.

%% file: myappendix.tex


\subsection{Scores}

\input{figures/tables/scores}

%% file: figures/tables/scores.tex
\begin{table}[!htb]
    \centering
    {\footnotesize
    \begin{tabular}{l|c c c c c c c c} \toprule
      \textbf{Encoder} & \textbf{BLEU} & \textbf{MOVER} & \textbf{PERM} & \textbf{Id} & \textbf{Id / PERM}\\
      \midrule
      Average & 33.51 & 67.62 & 43.22 & 15.18 & 35.12\\
      GEM & 23.76 & 28.58 & 1.44 & 1.21 & 84.03\\
      Hier & 37.9 & 58.61 & 17.29 & 14.97 & 86.58\\
      Avg+Max+Hier & 34.91 & 62.48 & 30.62 & 17.67 & 50.85\\
      Sent2Vec & 32.94 & 45.48 & 16.08 & 2.77 & 87.38\\
      \midrule
      Infersent & 53.54 & 72.01 & 23.21 & 21.19 & 91.29\\
      QuickThought & 48.07 & 70.09 & 22.27 & 20.02 & 89.9\\
      LASER & 52.88 & 73.87 & 33.58 & 18.52 & 96.51\\
      SBERT & 21.5 & 43.39 & 3.45 & 3.27 & 94.78\\
      \bottomrule
    \end{tabular}
    }
    \caption{Complete scores from the metrics evaluated on generated sentences. Experimental setup detailed in section \ref{sec:experiments}}
\end{table}

\begin{table}[!htb]
   \centering
   {\footnotesize
    \begin{tabular}{l|ccccc|cccccccccc} \toprule
        \textbf{Encoder} & \textbf{Average} & \textbf{GEM} & \textbf{Hier} & \textbf{Avg+Max+Hier} & \textbf{Sent2Vec} & \textbf{Infersent} & \textbf{Quickthoughts} & \textbf{SBERT}\\
        \midrule
        STS & 0.57 & 0.59 & 0.58 & 0.63 & 0.57 & 0.71 & 0.61 & 0.76\\
        MR & 72.18 & 63.46 & 68.34 & 72.16 & 75.16 & 79.44 & 82.58 & 84.79\\
        CR & 74.49 & 72.56 & 72.66 & 75.6 & 77.51 & 84.32 & 84.66 & 90.81\\
        MPQA & 74.83 & 73.26 & 74.48 & 76.19 & 87.44 & 89.37 & 89.97 & 90.43\\
        SUBJ & 89.47 & 78.24 & 86.53 & 89.16 & 91.58 & 92.64 & 94.8 & 94.47\\
        SST2 & 76.39 & 64.52 & 72.32 & 74.68 & 78.91 & 84.57 & 88.08 & 90.83\\
        SST5 & 39.1 & 32.71 & 37.65 & 39.1 & 40.68 & 45.79 & 48.64 & 50.09\\
        TREC & 69.6 & 66 & 64.4 & 69 & 76.4 & 90.6 & 91.4 & 86.6\\
        MRPC F1 & 80.95 & 72.31 & 79.74 & 81.75 & 80.74 & 83.74 & 84.05 & 82.87\\
        SICK E & 74.28 & 70.25 & 72.97 & 76.05 & 78.77 & 85.59 & 82.73 & 83.05\\
        SICK R & 0.63 & 0.57 & 0.59 & 0.67 & 0.72 & 0.83 & 0.81 & 0.8\\
        STSb & 0.57 & 0.47 & 0.55 & 0.61 & 0.53 & 0.78 & 0.79 & 0.75\\
        COCO & 26.1 & 16.32 & 21.24 & 28.8 & 29.01 & 43.02 & 42.32 & 38.12\\
        SNLI & 63.86 & 52.46 & 62.33 & 67.35 & 59.84 & 84.47 & 78.06 & 83.92\\
        \bottomrule
    \end{tabular}
    }
    \caption{Complete scores from downstream tasks evaluated using senteval.}
\end{table}